# A Momentum-Incorporated Non-Negative Latent Factorization of Tensors Model for Dynamic Network Representation

Aoling Zang

*Abstract*—A large-scale dynamic network (LDN) is a source of data in many big data-related applications due to their large number of entities and large-scale dynamic interactions. They can be modeled as a high-dimensional incomplete (HDI) tensor that contains a wealth of knowledge about time patterns. A Latent factorization of tensors (LFT) model efficiently extracts this time pattern, which can be established using stochastic gradient descent (SGD) solvers. However, LFT models based on SGD are often limited by training schemes and have poor tail convergence. To solve this problem, this paper proposes a novel nonlinear LFT model (MNNL) based on momentum-incorporated SGD, which extracts non-negative latent factors from HDI tensors to make training unconstrained and compatible with general training schemes, while improving convergence accuracy and speed. Empirical studies on two LDN datasets show that compared to existing models, the MNNL model has higher prediction accuracy and convergence speed.

## I. Introduction

Large-scale dynamic network (LDN) contains rich information about various expected patterns [1]-[3], often appearing in big data-related fields like social network services, financial analysis, etc., but there are many unobserved missing items. The identification and prediction of missing items, as a hot research field in complex networks [4]-[7], helps us understand the internal structural characteristics of many practical systems and helps us solve a series of important problems in natural and social systems, and has important theoretical Significance and practical application value. However, considering large-scale dynamic networks such as a telecommunication network, since the number of nodes is increasing, it becomes impossible to observe the complete interaction between nodes in each time slot [8]-[9]. Correspondingly, a LDN can be modeled as a typical high-dimensional incomplete tensor. Therefore, how to obtain rich knowledge from a HDI tensor becomes a critical and necessary problem.

According to previous studies [10]-[11], a latent factorization of tensors model can effectively extract useful knowledge from a HDI tensor [12]-[17]. However, the current LFT-based models are often linear [18]-[28], in other words, the LFT-based model does not consider the nonlinear mode in the HDI tensor. According to previous research [29]-[31], the neural network has the ability to characterize the nonlinear characteristics of the data, so the main idea of this paper is to integrate the idea of neural network and more accurately fit the HDI tensor when building the LFT model.

In order to solve the problem of insufficient expressive ability of the above-mentioned linear model, this paper will introduce the activation function to add nonlinear factors. The activation function is a function used in the neural network to calculate the weighted sum of the input and deviation, and is used to determine whether the neuron can be released. It is very important for the artificial neural network model to learn and understand very complex and nonlinear functions. They introduce nonlinear properties into our network. In addition, LFT-based models are usually effectively established by building an SGD solver [32]-[37], but SGD-based LFT often has low tail convergence, that is, it may fall into a local optimum during the gradient descent process, without considering the impact of the previous update process, resulting in slower convergence. Therefore, we extend SGD in this paper and introduce the method of combining momentum for optimization.

In overview, this paper has three main contributions as follows:

1) We propose a model based on CP decomposition to efficiently represent a LDN, which greatly improves the convergence speed of the proposed model by utilizing a stochastic gradient descent optimization method combined with momentum methods.

2) We introduce an activation function in this model to map the hidden feature factor matrix, which improves the nonlinearity of the neural network model.

Experiments on four real complex network datasets show that the proposed model is superior to the existing model in prediction accuracy and number of iteration rounds.

## II. Preliminaries

*A. HDI Tensor*

As shown in Figure 1, we use the LDN dataset as the input data source, and since the LDN dataset contains only a few known elements, most of which are unknown, the LDN dataset can usually be defined with an HDI tensor, as follows:

**Definition 1:** (HDI tensor): Let $I, J, K$ denote three entity sets, and $\mathbf{A}^{|I| \times |J| \times |K|}$ denote the target tensor where each element $a_{ijk}$ describes some relationship between node $i \in I$ to $j \in J$ at the time point $k \in K$. As $I, J$ and $K$ can be huge in many big-data applications, it becomes impossible to observe the full relationship among their entities. Consequently, **A** is usually an HDI tensor.

---

✧ A. Zang is the School of Computer and Information Science, Southwest University, Chongqing 400715, China (e-mail: zengaoll@163.com).

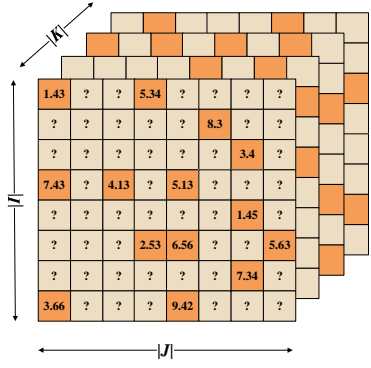

Fig. 1. An HDI tensor **A**

*B. Problem Formulation*

Let Λ and Γ represent the set of known and unknown entries in **A**, respectively, and we seek to the $(|I|+|J|+|K|)\times R$-dimensional latent factor matrix $X$, where $R$ represents the dimension of $X$, the specified entities $i \in I$, $j \in J$ and $k \in K$ correspond to vectors $x_{(i)}, x_{(j)}$ or $x_{(k)}$, respectively. In order to extract $X$ from a known set, an objective function is usually required, and in this paper, the problem is represented by using Euclidean distance. Therefore, we formulate this issue as follows:

$$\arg\min \varepsilon(X) = \frac{1}{2} \sum_{a_{i,j,k} \in \Lambda} \left( \left( a_{i,j,k} - \sum_{r=1}^{R} x_{(i)r} x_{(j)r} x_{(k)r} \right)^2 \right), \quad (1)$$

$$s.t. \forall i \in I, j \in J, k \in K, r \in \{1,2,\ldots,R\}: x_{(i)} \geq 0, x_{(j)} \geq 0, x_{(k)} \geq 0$$

### III. OUR MODEL

*A. Unconstrained Problem Formulation*

In question (1), because $X$ plays a dual role as both outputting the latent factor and decision parameter, it may lead to incompatibility with conventional learning schemes. For example, since the problem of non-negative constraints is not compatible with SGD. So, if we: 1) introduce an additional parameter to play the role of a decision parameter, problem (1) can be greatly simplified; 2) Keep $X$ as the output LF vector; 3) $X$ is constantly satisfied with non-negative constraints by a mapping function that concatenates $X$ and newly introduced parameters. Therefore, we introduce $(|I|+|J|+|K|)\times R$-dimensional vector $Y$ as decision arguments, and the single-element correlation mapping function $f(*)$, which maps each element in $Y$ to the corresponding element in $X$ to relax the non-negative constraint. The relationship between $X$ and $Y$ can be expressed as:

$$\forall i \in I, j \in J, k \in K, r \in \{1,2,3,\ldots,R\}:$$
$$x_{(i)r} = f\left(y_{(i)r}\right), x_{(j)r} = f\left(y_{(j)r}\right), x_{(k)r} = f\left(y_{(k)r}\right) \quad (2)$$

Since $x_{(i)}, x_{(j)}$ and $x_{(k)}$ are all non-negative latent factor matrices, we reformulate (1) as follows by making rule f satisfy $\forall y \in \mathbb{R}$, $f(y) > 0$:

$$\arg\min \varepsilon(X) = \frac{1}{2} \sum_{y_{i,j,k} \in \Lambda} \left( a_{i,j,k} - \sum_{r=1}^{R} f\left(y_{(i)r}\right) f\left(y_{(j)r}\right) f\left(y_{(k)r}\right) \right)^2 \quad (3)$$

where $X$ essentially outputs a potential factor that is non-negative. To make the solution spaces of (3) and (1) close and f not locally redundant, we select the activation function sigmoid, denoted $\Phi(\alpha)$, whose function expression is: $\Phi(\alpha) = (1+e^{-\alpha})^{-1}$ to map $Y$ to $X$:

$$\forall i \in I, j \in J, k \in K, r \in \{1,2,3,\ldots,R\}:$$
$$x_{(i)r} = \Phi\left(y_{(i)r}\right), x_{(j)r} = \Phi\left(y_{(j)r}\right), x_{(k)r} = \Phi\left(y_{(k)r}\right) \quad (4)$$

Bringing formula (4) into (3), we get the objective function:

$$\arg\min \varepsilon(Y) = \frac{1}{2} \sum_{a_{i,j,k} \in \Lambda} \left( a_{i,j,k} - \sum_{r=1}^{R} \Phi\left(y_{(i)r}\right) \Phi\left(y_{(j)r}\right) \Phi\left(y_{(k)r}\right) \right)^2 \quad (5)$$

As in previous studies, it is critical that we apply Tikhonov regularization to LFT models to prevent overfitting. By doing so, we extend (5) to:

$$\arg\min \varepsilon(Y) = \frac{1}{2} \sum_{a_{i,j,k} \in \Lambda} \left( \left( a_{i,j,k} - \sum_{r=1}^{R} \Phi\left(y_{(i)r}\right) \Phi\left(y_{(j)r}\right) \Phi\left(y_{(k)r}\right) \right)^2 + \lambda \sum_{r=1}^{R} \left( \Phi^2\left(y_{(i)r}\right) + \Phi^2\left(y_{(j)r}\right) + \Phi^2\left(y_{(k)r}\right) \right) \right) \quad (6)$$

## B. Momentum-incorporated SGD Solver

As demonstrated by previous studies, optimization with respect to Y can be achieved by most general learning schemes. SGD has the advantages of fast convergence and ease of implementation when performing LFT analysis of HDI tensor, so that (6) is minimized as follows:

$$\arg\min_{y_{(i)}, y_{(j)}, y_{(k)}} \varepsilon(y_{(i)}, y_{(j)}, y_{(k)}) \overset{SGD}{\Rightarrow} \forall i \in I, j \in J, k \in K, r \in \{1, 2, \ldots, R\}:$$

$$\begin{cases} y_{(i)r}^{t+1} \leftarrow y_{(i)r}^t - \eta \dfrac{\partial \varepsilon_{ijk}^t}{\partial x_{(i)r}^t} \dfrac{dx_{(i)r}^t}{dy_{(i)r}^t} \\ y_{(j)r}^{t+1} \leftarrow y_{(j)r}^t - \eta \dfrac{\partial \varepsilon_{ijk}^t}{\partial x_{(j)r}^t} \dfrac{dx_{(j)r}^t}{dy_{(j)r}^t} \\ y_{(k)r}^{t+1} \leftarrow y_{(k)r}^t - \eta \dfrac{\partial \varepsilon_{ijk}^t}{\partial x_{(k)r}^t} \dfrac{dx_{(k)r}^t}{dy_{(k)r}^t} \end{cases} \quad (7)$$

where $\varepsilon_{ijk}$ and $\eta$ represent the loss and step size on the training instance $a_{i,j,k} \in \Lambda$, respectively. A momentum approach that records the early iteration process and influences the direction of the next iteration. Given the parameter $\theta$ of target $J(\theta)$, gradient descent with momentum is updated as follows:

$$\begin{aligned} v_0 &= 0, \\ v_t &= \gamma v_{t-1} + \eta \nabla_\theta J(\theta_{t-1}), \\ \theta_t &= \theta_{t-1} - v_t. \end{aligned} \quad (8)$$

where $v_0$ is the velocity of the initial state, and $v_{t-1}$ and $v_t$ are update velocity vectors for iteration $(t-1)$ and $t$, respectively. $(t-1)$ and $t$ represent the state at the $t$ th and $(t-1)$ th iterations, respectively.

The idea of the MNNL method is to extend the base SGD in the LFT and combine it with the momentum generation method. Specifically, combining Equation (8) with Equation (7), we derive the update of $y_{(i)r}$ as follows:

$$v_{y_{(i)r}}^0 = 0, v_{y_{(i)r}}^{t+1} = \gamma v_{y_{(i)r}}^t + \eta \dfrac{\partial \varepsilon_{ijk}^t}{\partial x_{(i)r}^t} \dfrac{dx_{(i)r}^t}{dy_{(i)r}^t}, y_{(i)r}^{t+1} = y_{(i)r}^t - v_{y_{(i)r}}^{t+1}. \quad (9)$$

where gamma is the constant that adjusts the influence of momentum, and v is the update velocity vector of the $t$+1st iteration. Similar to $y_{(i)r}$, we infer that the updates for $y_{(j)r}$ and $y_{(k)r}$ are as follows:

$$v_{y_{(j)r}}^0 = 0, v_{y_{(j)r}}^{t+1} = \gamma v_{y_{(j)r}}^t + \eta \dfrac{\partial \varepsilon_{ijk}^t}{\partial x_{(j)r}^t} \dfrac{dx_{(j)r}^t}{dy_{(j)r}^t}, y_{(j)r}^{t+1} = y_{(j)r}^t - v_{y_{(j)r}}^{t+1} \quad (10)$$

$$v_{y_{(k)r}}^0 = 0, v_{y_{(k)r}}^{t+1} = \gamma v_{y_{(k)r}}^t + \eta \dfrac{\partial \varepsilon_{ijk}^t}{\partial x_{(k)r}^t} \dfrac{dx_{(k)r}^t}{dy_{(k)r}^t}, y_{(k)r}^{t+1} = y_{(k)r}^t - v_{y_{(k)r}}^{t+1} \quad (11)$$

Due to $\Phi'(\alpha) = \Phi(\alpha)(1 - \Phi(\alpha))$, and by making $e_{i,j,k} = a_{i,j,k} - \sum_{r=1}^{R} \Phi(y_{(i)r}) \Phi(y_{(i)r}) \Phi(y_{(j)r})$, bringing it into the equations (9), (10), (11), we get:

$$\arg\min_{y_{(i)r}, y_{(j)r}, y_{(k)r}, v_{y_{(i)r}}, v_{y_{(j)r}}, v_{y_{(k)r}}} \varepsilon (y_{(i)r}, y_{(j)r}, y_{(k)r}, v_{y_{(i)r}}, v_{y_{(j)r}}, v_{y_{(k)r}}) \overset{MSGD}{\Rightarrow} \forall i \in I, j \in J, k \in K, r \in \{1, 2, \ldots, R\}:$$

$$\begin{cases} v_{y_{(i)r}}^{t+1} = \gamma v_{y_{(i)r}}^t + \eta \Phi(y_{(i)r}^t)(1 - \Phi(y_{(i)r}^t))(\Phi(y_{(j)r}^t) \Phi(y_{(k)r}^t) e_{i,j,k} - \lambda(y_{(i)r}^t)) \\ v_{y_{(j)r}}^{t+1} = \gamma v_{y_{(j)r}}^t + \eta \Phi(y_{(j)r}^t)(1 - \Phi(y_{(j)r}^t))(\Phi(y_{(i)r}^t) \Phi(y_{(k)r}^t) e_{i,j,k} - \lambda(y_{(j)r}^t)) \\ v_{y_{(k)r}}^{t+1} = \gamma v_{y_{(k)r}}^t + \eta \Phi(y_{(k)r}^t)(1 - \Phi(y_{(k)r}^t))(\Phi(y_{(i)r}^t) \Phi(y_{(j)r}^t) e_{i,j,k} - \lambda(y_{(k)r}^t)) \\ y_{(i)r}^{t+1} = y_{(i)r}^t - v_{y_{(i)r}}^{t+1} \\ y_{(j)r}^{t+1} = y_{(j)r}^t - v_{y_{(j)r}}^{t+1} \\ y_{(k)r}^{t+1} = y_{(k)r}^t - v_{y_{(k)r}}^{t+1} \end{cases} \quad (12)$$

## IV. EXPERIMENTAL RESULTS AND DISCUSSION

### A. General Settings

*1) Datasets:* We conducted experiments with two real LDN datasets, which involved two real Bitcoin datasets. Note that for beneficial purposes, all node IDs are encrypted and the data is normalized. Based on the above principles, two HDI tensors are built, the details of which are shown in Table 1. For each dataset, we randomly divide it into training set, validation set and test set, we perform multiple experiments on each dataset, and take the average of multiple experimental results to eliminate accidents.

TABLE I. Dataset Details

| Datasets | Nodes | Time Points | Entries | Density |
|---|---|---|---|---|
| **D1** | 7604 | 165 | 24186 | $2.53 \times 10^{-6}$ |
| **D2** | 6005 | 165 | 35592 | $5.98 \times 10^{-6}$ |

*2) Evaluation Metrics:* In this paper, we choose RMSE to reflect the prediction accuracy of the model on the missing data of the HDI tensor. Lower RMSE indicates higher prediction accuracy of the model for tensor missing data. If $\hat{a}_{ijk}$ and $a_{ijk}$ are estimated and actual, respectively, the two expressions can be written as follows.

$$\text{RMSE} = \sqrt{\frac{\sum_{a_{ijk} \in \Lambda} (a_{ijk} - \hat{a}_{ijk})^2}{|\Lambda|}}$$

### B. Compared Models

In the next, we compare our model with the following two advanced models.
M1: A multi-dimensional tensor model in [40]. It adopts CPD framework to build the LFT model, and alternatively trains the desired LFs via alternating least square (ALS) and gradient descent algorithms.
M2: A biased nonnegative tensor factorization model in [41]. It incorporates linear biases into the model for describing QoS fluctuations, and it adds nonnegative constraints to the factor matrices as well.
M3: The model proposed in this paper.

### C. Experimental Results

The experimental results of M1-3 are shown in Table 2. From these results, it can be seen that the proposed model M3 has higher prediction accuracy than other similar algorithms, and the convergence speed is also improving. As shown in Table 2, M3 outperforms other models in predicting different network data. For example, M3 has a minimum RMSE of 0.4734 on D1 and 98 iteration counts when the minimum RMSE is reached. The other two models have a minimum RMSE of 0.4929 and 0.4826 on D1, respectively. The iteration counts at the minimum RMSE is 497 and 124, respectively. Therefore, the prediction accuracy of model M3 on D1 is 3.07% and 1.74% higher than model M1 and model M2, respectively, and the iteration counts is reduced by 399 and 26, respectively. Similar results can be obtained on dataset D2.

TABLE II. Lowest RMSE and MAE of Each Model on All Testing Case.

| Datasets | M1 | M2 | M3 |
|---|---|---|---|
| **D1** | 0.4929/497 | 0.4826/124 | **0.4734/98** |
| **D2** | 0.5232/250 | 0.5048/96 | **0.4793/24** |

## V. CONCLUSION

In this paper, we propose an LFT model based on the CP decomposition framework, which uses a nonlinear activation function to map potential factors, adding a momentum method to stochastic gradient descent. Finally, experiments are carried out on two real network datasets and compared with existing models, and the experiments show that our proposed model has higher prediction accuracy.

REFERENCES

[1] S. Li, M. Zhou, X. Luo and Z. You, "Distributed Winner-Take-All in Dynamic Networks," *IEEE Transactions on Automatic Control*, vol. 62, no. 2, pp. 577-589, 2017.
[2] Y. Han, G. Huang, S. Song, L. Yang, H. Wang, and Y. Wang. "Dynamic neural networks: A survey," *IEEE Transactions on Pattern Analysis and Machine Intelligence*, vol. 44, no.10, pp. 7436-7456, 2021.
[3] S. Li, M. Zhou and X. Luo, "Modified Primal-Dual Neural Networks for Motion Control of Redundant Manipulators with Dynamic Rejection of Harmonic Noises," *IEEE Transactions on Neural Networks and Learning Systems*, vol. 29, no. 10, pp. 4791-4801, 2018.
[4] J. Jiang, and Y. Lai. "Model-free prediction of spatiotemporal dynamical systems with recurrent neural networks: Role of network spectral radius," *Physical Review Research*, vol. 1, no. 3, pp. 033-056, 2019.
[5] Q. Xuan, Z. Zhang, C. Fu, H. Hu, and V. Filkov, "Social synchrony on complex networks," *IEEE Trans. on Cybernetic*, vol. 48, no. 5, pp. 1420–1431, 2018.

[6] X. Luo, H. Wu, and Z. Li, "NeuLFT: A Novel Approach to Nonlinear Canonical Polyadic Decomposition on High-Dimensional Incomplete Tensors," *IEEE Transactions on Knowledge and Data Engineering*, DOI: 10.1109/TKDE.2022.3176466.

[7] H. Wu, and X. Luo. "Instance-Frequency-Weighted Regularized, Nonnegative and Adaptive Latent Factorization of Tensors for Dynamic QoS Analysis," *In Proc. of the 2021 IEEE Int. Conf. on Web Services. (ICWS2021) (Regular)*, Chicago, IL, USA , 2021, pp. 560-568.

[8] X. Luo, Y. Zhou, Z. Liu, and M. Zhou, "Fast and Accurate Non-negative Latent Factor Analysis on High-dimensional and Sparse Matrices in Recommender Systems," *IEEE Transactions on Knowledge and Data Engineering*, DOI: 10.1109/TKDE.2021.3125252.

[9] D. Wu, and X. Luo, "Robust Latent Factor Analysis for Precise Representation of High-dimensional and Sparse Data", *IEEE/CAA Journal of Automatica Sinica*, vol. 8, no. 4, pp. 796-805, 2021.

[10] X. Luo, H. Wu, Z. Wang, J. Wang, and D. Meng, "A Novel Approach to Large-Scale Dynamically Weighted Directed Network Representation," *IEEE Transactions on Pattern Analysis and Machine Intelligence*, vol. 44, no. 12, pp. 9756-9773, 2022.

[11] M. Chen, C. He, and X. Luo. "MNL: A Highly-Efficient Model for Large-scale Dynamic Weighted Directed Network Representation," *IEEE Transactions on Big Data*, DOI: 10.1109/TBDATA.2022.3218064.

[12] H. Li, P. Wu, N. Zeng, Y. Liu, and F. Alsaadi, "A Survey on Parameter Identification, State Estimation and Data Analytics for Lateral Flow Immunoassay: from Systems Science Perspective," *International Journal of Systems Science*, DOI:10.1080/00207721.2022.2083262.

[13] D. Wu, X. Luo, M. Shang, Y. He, G. Wang, and X. Wu, "A Data-Characteristic-Aware Latent Factor Model for Web Services QoS Prediction," *IEEE Transactions on Knowledge and Data Engineering*, vol. 34, no. 6, pp. 2525-2538, 2022.

[14] Z. Liu, G. Yuan, and X. Luo, "Symmetry and Nonnegativity-Constrained Matrix Factorization for Community Detection," *IEEE/CAA Journal of Automatica Sinica*, vol. 9, no. 9, pp. 1691-1693, 2022.

[15] H. Wu, X. Luo, and M. C. Zhou. "Neural Latent Factorization of Tensors for Dynamically Weighted Directed Networks Analysis," *In Proc. of the 2021 IEEE Int. Conf. on Systems, Man, and Cybernetics*, Melbourne, Australia, 2021, pp. 3061-3066.

[16] X. Luo, Y. Yuan, S. Chen, N. Zeng, and Z. Wang, "Position-Transitional Particle Swarm Optimization-Incorporated Latent Factor Analysis," *IEEE Transactions on Knowledge and Data Engineering*, vol. 34, no. 8, pp. 3958-3970, 2022.

[17] J. Chen, X. Luo, and M. Zhou, "Hierarchical Particle Swarm Optimization-incorporated Latent Factor Analysis for Large-Scale Incomplete Matrices," *IEEE Transactions on Big Data*, vol. 8, no. 6, pp. 1524-1536, 2022.

[18] X. Luo, M. Zhou, S. Li, L. Hu, and M. Shang, "Non-negativity Constrained Missing Data Estimation for High-dimensional and Sparse Matrices from Industrial Applications," *IEEE Transactions on Cybernetics*, vol. 50, no. 5, pp. 1844-1855, 2020.

[19] W. Zhang, H. Sun, X. Liu, and X. Guo, "Temporal QoS-aware web service recommendation via non-negative tensor factorization," in *Proceedings of the 23rd international conference on World wide web*, New York, United States, 2014, pp. 585-596.

[20] X. Luo, H. Wu, M. Zhou and H. Yuan, "Temporal Pattern-aware QoS Prediction via Biased Non-negative Latent Factorization of Tensors," *IEEE Transactions on Cybernetics*, vol. 50 , no. 5, pp. 1798-1809, 2020.

[21] H. Wu, X. Luo, and M. Zhou, "Advancing Non-negative Latent Factorization of Tensors with Diversified Regularizations," *IEEE Transactions on Services Computing*, vol. 15, no. 3, pp. 1334-1344, 2022.

[22] X. Luo, M. Chen, H. Wu, Z. Liu, H. Yuan, and M. Zhou, "Adjusting Learning Depth in Non-negative Latent Factorization of Tensors for Accurately Modeling Temporal Patterns in Dynamic QoS Data," *IEEE Transactions on Automation Science and Engineering*, vol. 18, no. 4, pp. 2142-2155, 2022.

[23] Z. Liu, X. Luo, and M. Zhou, "Symmetry and Graph Bi-regularized Non-Negative Matrix Factorization for Precise Community Detection," *IEEE Transactions on Automation Science and Engineering*, DOI: 10.1109/TASE.2023.3240335.

[24] H. Wu, X. Luo, M. C. Zhou, M. J. Rawa, K. Sedraoui, and A. Albeshri. "A PID-Incorporated Latent Factorization of Tensors Approach to Dynamically Weighted Directed Network Analysis," *IEEE/CAA Journal of Automatica Sinica*, vol. 9, no. 3, pp.533-546, 2022.

[25] H. Wu, X. Luo, and M. C. Zhou. "Discovering Hidden Pattern in Large-scale Dynamically Weighted Directed Network via Latent Factorization of Tensors," *In Proc. of the 17th IEEE Int. Conf. on Automation Science and Engineerin*, Lyon, France, 2021, pp. 1533-1538.

[26] Y. Song, Z. Zhu, M. Li, G. Yang, and X. Luo, "Non-negative Latent Factor Analysis-Incorporated and Feature-Weighted Fuzzy Double c-Means Clustering for Incomplete Data," *IEEE Transactions on Fuzzy Systems*, vol. 30, no. 10, pp. 4165-4176, 2022.

[27] X. Luo, M. Zhou, S. Li, and M. Shang, "An Inherently Non-negative Latent Factor Model for High-dimensional and Sparse Matrices from Industrial Applications," *IEEE Transactions on Industrial Informatics*, vol. 14, no. 5, pp. 2011-2022, 2018.

[28] Z. S Lin, and H. Wu. "Dynamical Representation Learning for Ethereum Transaction Network via Non-negative Adaptive Latent Factorization of Tensors," *In Proc. of the 2021 Int. Conf. on Cyber-physical Social Intelligence*, Beijing, China, pp. 1-6, 2021.

[29] X. Luo, J. Sun, Z. Wang, S. Li, and M. Shang, "Symmetric and Non-negative Latent Factor Models for Undirected, High Dimensional and Sparse Networks in Industrial Applications," *IEEE Transactions on Industrial Informatics*, vol. 13, no. 6, pp. 3098-3107, 2017.

[30] W. Li, Q. He, X. Luo, and Z. Wang, "Assimilating Second-Order Information for Building Non-Negative Latent Factor Analysis-Based Recommenders," *IEEE Transactions on System Man Cybernetics: Systems*, vol. 52, no. 1, pp. 485-497, 2021.

[31] X. Luo, M. Zhou, Y. Xia, Q. Zhu, A. Ammari, and A. Alabdulwahab, "Generating Highly Accurate Predictions for Missing QoS-data via Aggregating Non-negative Latent Factor Models," *IEEE Transactions on Neural Networks and Learning Systems*, vol. 27, no. 3, pp. 579-592, 2016.

[32] Y. Xu, Z. Wu, J. Chanussot, and Z. Wei. "Hyperspectral images super-resolution via learning high-order coupled tensor ring representation," *IEEE transactions on neural networks and learning systems*, vol. 31, no. 11, pp. 4747-4760, 2020.

[33] Z. Long, C. Zhu, J. Liu, and Y. Liu. "Bayesian low rank tensor ring for image recovery," *IEEE Transactions on Image Processing*, vol. *30*, pp. 3568-3580, 2021.

[34] Q. Zhao, G. Zhou, S. Xie, L. Zhang, and A. Cichocki. "Tensor ring decomposition," *arXiv preprint arXiv:1606.05535*, 2016.

[35] W. He, Y. Chen, N. Yokoya, C. Li, and Q. Zhao. "Hyperspectral super-resolution via coupled tensor ring factorization," *Pattern Recognition*, 122, 108280, 2022.

[36] J. Yu, G. Zhou, W. Sun, and S. Xie. "Robust to rank selection: Low-rank sparse tensor-ring completion," *IEEE Transactions on Neural Networks and Learning Systems*, DOI: 10.1109/TNNLS.2021.3106654.

[37] X. Luo, Y. Zhou, Z. Liu, L. Hu, and M. Zhou, "Generalized Nesterov's Acceleration-incorporated, Non-negative and Adaptive Latent Factor Analysis," *IEEE Transactions on Services Computing*, vol. 15, no. 5, pp. 2809-2823, 2022.

[38] N. Zeng, P. Wu, Z. Wang, H. Li, W. Liu, X. Liu, "A small-sized object detection oriented multi-scale feature fusion approach with application to defect detection," *IEEE Transactions on Instrumentation and Measurement*, vol. 71, no. 35, PP. 7-14, 2022.

[39] Xing. Su, M. Zhang, Y. Liang, Z. Cai, L. Guo, and Z. Ding, "A tensor-based approach for the QoS evaluation in service-oriented environments," *IEEE Transactions on Network and Service Management*, vol. 18, no. 3, pp. 3843-3857, 2021.

[40] S. Wang, Y. Ma, B. Cheng, F. Yang, and R. N. Chang, "Multi-dimensional QoS prediction for service recommendations," *IEEE Transactions on Services Computing*, vol. 12, no.1, pp. 47-57, 2019.